\documentclass[]{research}

\usepackage[utf8]{inputenc}
\usepackage[T1]{fontenc}
\usepackage{url}
\usepackage{amsfonts}
\usepackage{amsmath}
\usepackage{mathtools}
\usepackage{amssymb}
\usepackage{nicefrac}
\usepackage{fp}
\usepackage{colortbl}
\definecolor{lightblue}{HTML}{DAEEF7}
\definecolor{lightred}{HTML}{FDDEDE}
\definecolor{colFC}{RGB}{132,173,241}   % FrontierCS  (paper(G_BLUE))
\definecolor{colFS}{RGB}{223,131,123}   % FrontierSmith (paper(G_RED))
\definecolor{colHT}{RGB}{241,201,91}    % HardTests  (paper(G_YELLOW))
\definecolor{colALE}{RGB}{196,135,206}  % ALE-bench  (paper(G_PURPLE))
\usepackage{graphicx}
\usepackage{subcaption}
\usepackage{wrapfig}
\usepackage[ruled,linesnumbered]{algorithm2e}
\SetKwInput{KwIn}{Input}
\SetKwInput{KwOut}{Output}
\SetKwComment{tmark}{$\triangleright$\ }{}
\usepackage{enumitem}
\usepackage{tikz}
\usepackage{pgfplots}
\pgfplotsset{compat=1.18}
\usepgfplotslibrary{groupplots}
\usepackage{pgfplotstable}
\usetikzlibrary{positioning, arrows.meta, calc, decorations.pathreplacing, shapes.geometric}
\usepackage{fontawesome5}
\usepackage{inconsolata}
\usepackage{xspace}
\DeclareRobustCommand{\hficon}{\raisebox{-0.18em}{\includegraphics[height=1.05em]{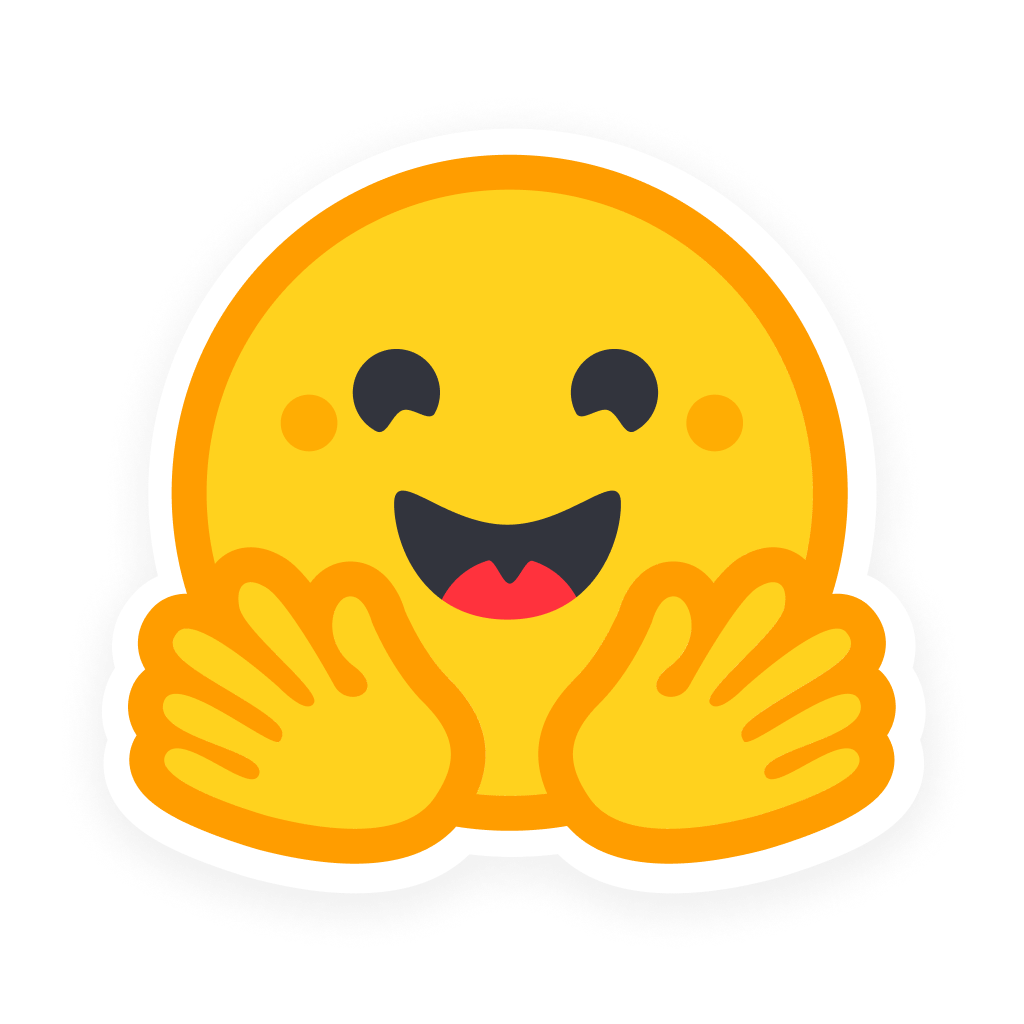}}}
\DeclareRobustCommand{\githubicon}{\faGithub}

\newcommand{\nBBaseFC}{1.80}
\newcommand{\nBBaseALE}{327.22}
\newcommand{\nBHumanFC}{11.17}
\newcommand{\nBHumanALE}{558.49}
\newcommand{\nBOursFC}{10.62}
\newcommand{\nBOursALE}{633.58}
\newcommand{\nBHardFC}{5.38}
\newcommand{\nBHardALE}{397.18}
\newcommand{\nBRandFC}{3.04}
\newcommand{\nBRandALE}{376.82}
\newcommand{\tBBaseFC}{7.70}
\newcommand{\tBBaseALE}{352.52}
\newcommand{\tBHumanFC}{13.98}
\newcommand{\tBHumanALE}{543.80}
\newcommand{\tBOursFC}{19.82}
\newcommand{\tBOursALE}{661.64}
\newcommand{\tBHardFC}{11.20}

\FPeval\nBDeltaFCval{round(\nBOursFC - \nBBaseFC, 2)}
\newcommand{\nBDeltaFC}{\nBDeltaFCval}
\FPeval\nBDeltaALEval{round(\nBOursALE - \nBBaseALE, 2)}
\newcommand{\nBDeltaALE}{\nBDeltaALEval}
\FPeval\tBDeltaFCval{round(\tBOursFC - \tBBaseFC, 2)}
\newcommand{\tBDeltaFC}{\tBDeltaFCval}
\FPeval\tBDeltaALEval{round(\tBOursALE - \tBBaseALE, 2)}
\newcommand{\tBDeltaALE}{\tBDeltaALEval}
\FPeval\nBGapHardFCval{round(\nBOursFC - \nBHardFC, 2)}
\newcommand{\nBGapHardFC}{\nBGapHardFCval}
\FPeval\nBGapHardALEval{round(\nBOursALE - \nBHardALE, 2)}
\newcommand{\nBGapHardALE}{\nBGapHardALEval}
\FPeval\nBGapRandFCval{round(\nBOursFC - \nBRandFC, 2)}
\newcommand{\nBGapRandFC}{\nBGapRandFCval}
\FPeval\nBGapRandALEval{round(\nBOursALE - \nBRandALE, 2)}
\newcommand{\nBGapRandALE}{\nBGapRandALEval}
\FPeval\nBGapHumanFCval{round(\nBHumanFC - \nBOursFC, 2)}
\newcommand{\nBGapHumanFC}{\nBGapHumanFCval}

\title{FrontierSmith: Synthesizing Open\nobreakdash-Ended Coding Problems at Scale}

\author[1,*]{Runyuan He}
\author[1,*,\S]{Qiuyang Mang}
\author[2]{Shang Zhou}
\author[3]{Kaiyuan Liu}
\author[1]{Hanchen Li}
\author[1]{Huanzhi Mao}
\author[4]{Qizheng Zhang}
\author{Zerui Li}
\author[5]{Bo Peng}
\author{Lufeng Cheng}
\author[6]{Tianfu Fu}
\author[1]{Yichuan Wang}
\author[5]{Wenhao Chai}
\author[2]{Jingbo Shang}
\author[1,7]{Alex Dimakis}
\author[1]{Joseph E. Gonzalez}
\author[1]{Alvin Cheung}

\affiliation[1]{UC Berkeley}
\affiliation[2]{UC San Diego}
\affiliation[3]{University of Washington}
\affiliation[4]{Stanford University}
\affiliation[5]{Princeton University}
\affiliation[6]{Massachusetts Institute of Technology}
\affiliation[7]{Bespoke Labs}

\contribution[*]{Equal contribution}
\contribution[\S]{Project lead}
\correspondence{\email{qmang@berkeley.edu}, \email{wenhao.chai@princeton.edu}}
\metadata[Model]{\hficon\ \href{https://huggingface.co/runyuanhe/qwen35-9b-frontiersmith}{huggingface.co/runyuanhe/qwen35-9b-frontiersmith}}
\metadata[Sample Problems \& Training Scripts]{\githubicon\ \href{https://github.com/FrontierCS/FrontierSmith}{github.com/FrontierCS/FrontierSmith}}

\abstract{
Many real-world coding challenges are open-ended and admit no known optimal solution. Yet, recent progress in LLM coding has focused on well-defined tasks such as feature implementation, bug fixing, and competitive programming. %\alvin{there are also multiple ways to fix a bug?} %\alvin{what does underexplored mean?}
Open-ended coding remains a weak spot for LLMs, largely because open-ended training problems are scarce and expensive to construct.
Our goal is to synthesize open-ended coding problems at scale to train stronger LLM coders.
We introduce \emph{FrontierSmith}, an automated system for iteratively evolving open-ended problems from existing closed-ended
coding tasks.
Starting from competitive programming problems, FrontierSmith generates candidate open-ended variants by changing the problems' goals, restricting outputs, and generalizing inputs.
It then uses a quantitative idea divergence metric to select problems that elicit genuinely diverse approaches from different solvers.
Agents then generate test cases and verifiers for the surviving candidates.
On two open-ended coding benchmarks, training on our synthesized data yields substantial gains over the base models: Qwen3.5-9B improves by +\nBDeltaFC{} score on FrontierCS and +\nBDeltaALE{} (Elo-rating-based performance) on ALE-bench; Qwen3.5-27B improves by +\tBDeltaFC{} and +\tBDeltaALE{}, respectively.
The synthesized problems also make agents take more turns and use more tokens, similar to human-curated ones, suggesting that closed-ended seeds can be a practical starting point for long-horizon coding data.
}

\begin{document}

\maketitle
\begingroup
\renewcommand{\thefootnote}{}
\footnotetext{*Equal contribution, \S Project lead}
\endgroup

\section{Introduction}

% Large language models (LLMs) now achieve strong performance on well-defined coding tasks.
% On competitive programming, models now reach gold-medal performance at ICPC~\citep{icpc};
% on software engineering, frontier models resolve over 80\% of real-world GitHub issues on SWE-bench Verified~\citep{swebench}.
% Yet these tasks share a common structure: correctness is well-defined and verifiable by binary test cases. By contrast, a coding task is \emph{open-ended} if no efficient method certifies its optimum at the scale of interest, and submissions are scored on a continuous quality scale. Open-endedness is a property of both the problem and its scale, not of computational complexity alone. SAT is NP-hard yet closed-ended; NP-hard optimization problems become open-ended once exact optima are out of reach in practice.
% On FrontierCS~\citep{frontiercs}, human experts score 95.41 on algorithmic tasks while frontier models like Gemini 3.0 Pro reach only 29.37. On ALE-bench~\citep{alebench}, frontier LLMs score below the average human participant in real heuristic contests.
% This gap reflects a fundamental asymmetry in training data: closed-ended tasks enjoy abundant problems and verified solutions, while open-ended tasks have almost none. Our goal is to automatically synthesize open-ended coding problems at scale to train stronger LLM coders.

LLMs now excel at well-defined coding tasks, reaching gold-medal performance in competitive programming~\citep{icpc} and solving over 80\% of SWE-bench verified issues~\citep{swebench}. Yet these settings are mostly closed-ended: correctness is discrete and efficiently verifiable. Open-ended coding tasks, in contrast, lack tractable certificates of optimality at the target scale and score submissions by continuous quality. 
For example, in cloud cluster scheduling, many job-to-machine assignments are feasible, but they differ continuously in makespan, tail latency, energy use, and utilization; checking feasibility is easy, whereas certifying global optimality at production scale is intractable~\cite{li2026skynomad}.
This difficulty is reflected in current open-ended benchmarks. On FrontierCS~\citep{frontiercs}, human experts score 95.41 on algorithmic tasks, compared with 29.37 for Gemini 3.0 Pro~\citep{google2025gemini3pro}; on ALE-bench~\citep{alebench}, frontier LLMs still trail average human participants in heuristic contests. This gap reflects a data asymmetry: verified closed-ended tasks are abundant, while open-ended tasks remain scarce. We aim to automatically synthesize open-ended coding problems at scale to train stronger LLM coders.

\begin{figure}
    \centering
    \makebox[\textwidth][c]{\includegraphics[width=1\textwidth]{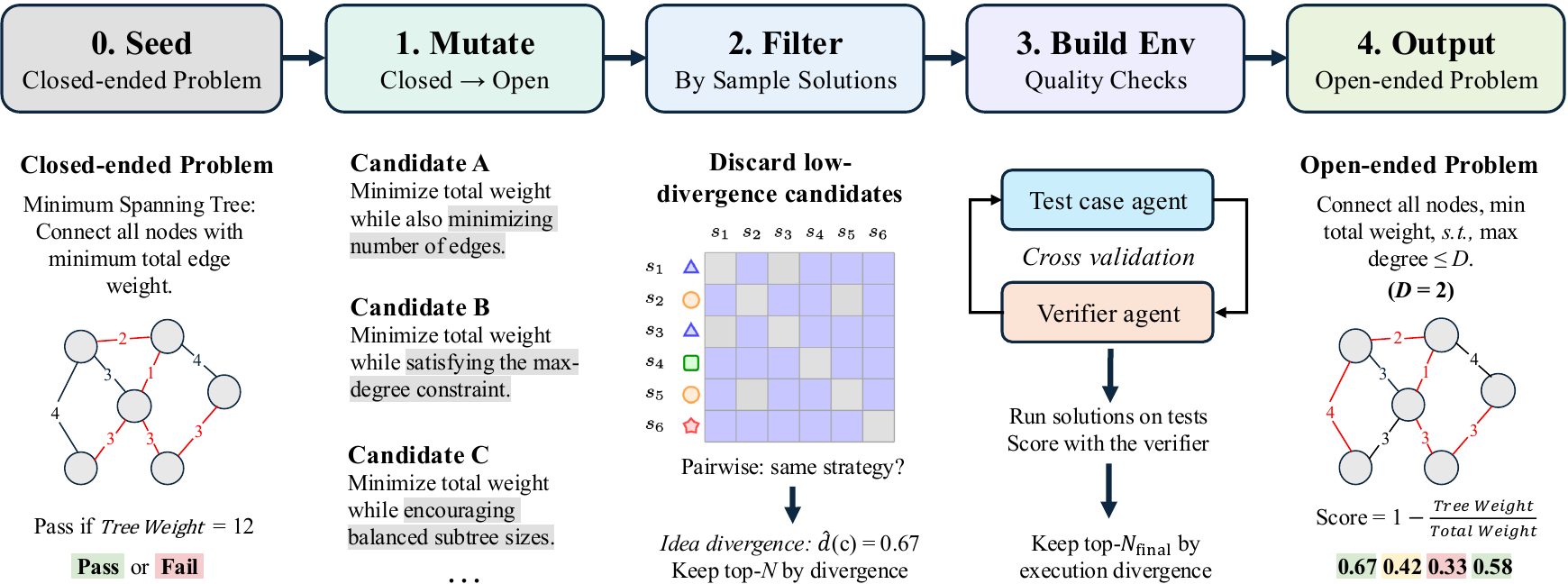}}
    \vspace{1ex}
    \caption{\textbf{\text{FrontierSmith} pipeline.} FrontierSmith converts closed-ended coding problems into open-ended ones through mutation, filtering, and automated environment construction. Candidate mugtations are ranked by idea divergence across sampled solutions, then validated by test-case and verifier agents. The resulting problems replace binary pass/fail feedback with continuous scores, enabling scalable training data synthesis for open-ended coding.}
    \label{fig:teaser}
\end{figure}

%\alvin{need to connect with previous para}
% \alvin{mention that our problems are also long horizon}
For closed-ended tasks, the data problem is largely solved. Codeforces~\citep{codeforces} and LeetCode~\citep{leetcode} provide hundreds of thousands of closed-ended problems with verified solutions, enabling reinforcement learning from verifiable rewards to drive substantial gains.
Open-ended coding has no equivalent.
FrontierCS and ALE-bench, two of the largest open-ended coding benchmarks, %\alvin{leading in what sense}
contains only around 240 and 40 human-curated %\alvin{algorithmic doesn't mean open ended, so why is this a problem?}
problems, respectively.

Constructing such %\alvin{algorithmic?}
problems manually is prohibitively expensive: each requires a carefully designed optimization objective, a verifier that produces continuous scores rather than binary judgments, and reliable test cases. %\alvin{closed ended problems require the same thing too? esp they want to assign partial credit rather than just 0/1}
Beyond engineering cost, expert judgment is needed to verify each problem is genuinely open-ended and not dominated by a single strategy (i.e., the core idea shared by a class of solutions, regardless of implementation). %\alvin{what does strategy mean?}

Recent work has automated coding data synthesis, but is exclusively for closed-ended, binary-correctness settings.
Given existing problem statements, several methods generate high-quality test cases at scale~\citep{hardtests, codecontestsplus, rstarcoder}.
Other methods synthesize new problems across competitive programming~\citep{autocode}, software engineering~\citep{swesmith, swerebench, bugpilot}, terminal environments~\citep{endlessterminals}, and self-play~\citep{absolutezero, gasp}.
None of them transfers to open-ended problems, which need novel formulations without a known optimum, and a way to reliably score solution quality on a continuous scale. %\alvin{again I don't see why this is unique to open ended problems}

We build \emph{FrontierSmith}, an automated system that continuously evolves open-ended coding problems from closed tasks at scale.
As shown in \Cref{fig:teaser}, starting from closed-end competitive programming problems, which provide a large and diverse seed corpus, %\alvin{why start from these? why not just start from existing open ended problems?}
FrontierSmith's pipeline mutates them into open-ended variants along three axes: changing goals, restricting outputs, and generalizing inputs. Each mutation turns a problem with a known efficient solution into one where exact solutions are intractable, forcing solvers to adopt diverse strategies. %\alvin{need a sentence to describe what these are and why doing so make them open ended}
We then quantify the open-endedness of the remaining candidates with a novel 
\emph{idea divergence} metric, which estimates the probability that two independent solvers use different core algorithms.
This estimation proceeds in two stages: first via LLM-based comparison of solution strategies, then via test-score similarity once test cases and verifiers are constructed for each candidate.
%\alvin{what is `that'? frontierSmith? the metric?}
These are built by separate agents systems that cross-validate each other to ensure correctness.
Validated problems then join the seed pool, so each subsequent round draws from an increasingly diverse starting set. %\alvin{I suggest structuring this para based on fig 1. Give a brief description of the overall process. Use the same terms}

We train Qwen3.5-9B and Qwen3.5-27B~\citep{qwen35blog} on 200 synthesized problems %\alvin{state how much data}
with GRPO~\citep{shao2024deepseekmath}. The 9B model gains +\nBDeltaFC{} score on FrontierCS and +\nBDeltaALE{} (Elo-rating-based performance) on ALE-bench; the 27B model gains +\tBDeltaFC{} and +\tBDeltaALE{}, respectively. %\alvin{grammar}
Against two controls, closed-ended HardTests~\citep{hardtests} and random rewards~\citep{shao2025spurious}, our synthetic problems lead by +\nBGapHardFC{} / +\nBGapHardALE{} and +\nBGapRandFC{} / +\nBGapRandALE{} on FrontierCS / ALE-bench, confirming that open-ended formulations and genuine reward signals are both necessary. Ablations show the filter is critical: removing it drops performance by 2.05 points on FrontierCS. Beyond training, our idea divergence metric cleanly separates real open-ended problems from competitive programming problems, validating it as a problem-quality classifier. %\alvin{what's the point of this sentence? you already described the results}
Finally, open-ended problems have recently shown to elicit distinctive long-horizon behavior in code agents, including more turns, tool calls, and thinking tokens~\cite{cemri2026adaevolve,liu2026evox, qu2026coral,autoevolver2026}; our synthetic problems match these patterns, suggesting they capture the same structure as human-curated ones.

In sum, our key findings are: (1) closed-ended problems can serve as seeds for open-ended synthesis via mutations that remove known optima; (2) idea divergence is an effective and tractable signal for selecting high-quality synthetic problems; and (3) FrontierSmith-generated problems yield training gains comparable to human-curated data, and exhibit long-horizon behavior when processed by agents.

\section{Related Work}
\label{sec:related}

\paragraph{Coding data synthesis for closed tasks.}
Recent work synthesizes coding problems and test infrastructure under binary correctness.
AutoCode~\citep{autocode}, SWE-smith~\citep{swesmith}, BugPilot~\citep{bugpilot}, Endless Terminals~\citep{endlessterminals}, GASP~\citep{gasp}, and SGS~\citep{sgs} synthesize problems;
HardTests~\citep{hardtests}, CodeContests+~\citep{codecontestsplus}, rStar-Coder~\citep{rstarcoder}, SWE-rebench~\citep{swerebench}, and R2E-Gym~\citep{r2egym} build tests and verifiers; AgentCoder~\citep{agentcoder} and CURE~\citep{cure} cross-validate code and tests across multiple agents.
None produces open-ended problems, the regime FrontierSmith targets. %\alvin{I thought you said the goal is to generate open ended problems}

\paragraph{Open-ended evaluation and solution diversity.}
Open-ended coding benchmarks score solutions %\alvin{what are outputs}
on a continuous quality scale: FrontierCS~\citep{frontiercs}, ALE-bench~\citep{alebench}, HeuriGym~\citep{heurigym}, KernelBench~\citep{kernelbench}, RE-Bench~\citep{rebench}, and MLE-bench~\citep{mlebench}.
NP-Engine~\citep{npengine} hand-crafts instance generators and rule-based verifiers for ten classical NP-hard tasks, training via RLVR with approximation-ratio rewards; its catalog is fixed and only difficulty varies, which caps the diversity of generated problems. Recent work~\citep{alphaevolve,wang2025thetaevolve,maheswaran2026squeeze} produces frontier results in open-ended mathematics and algorithm design.
Our idea divergence metric draws on quality-diversity and novelty search~\citep{lehman2011novelty, mouret2015mapelites, wang2019poet, bradley2023qdaif, faldor2024omni, aces}, with code-specific analogs in \citet{lee2025algodiv} and \citet{ju2025rpd}.
Unlike these works, which measure diversity of one model's outputs, we use inter-solver divergence as a problem-quality filter that admits only formulations capable of eliciting different core algorithms, a necessary but not sufficient signal for open-endedness. %\alvin{that doesn't mean the problem is open ended right? there can be multiple algos to fix a bug}

\paragraph{Mutation-based synthesis and iterative self-play.}
Evolving prompts or programs by mutation is established: WizardLM~\citep{wizardlm}, WizardCoder~\citep{wizardcoder}, EvoEval~\citep{evoeval}, and Auto Evol-Instruct~\citep{autoevol} mutate seed prompts; FunSearch~\citep{funsearch}, AlphaEvolve~\citep{alphaevolve}, ELM~\citep{lehman2022elm}, and EoH~\citep{eoh} mutate programs inside evolutionary loops.
Iterative bootstrap recycles self-generated data through training rounds: STaR~\citep{star}, ReST-EM~\citep{restem}, V-STaR~\citep{vstar}, Self-Rewarding LM~\citep{selfrewarding}, R-Zero~\citep{rzero}, EVA~\citep{eva}, and Absolute Zero~\citep{absolutezero}.
FrontierSmith differs: we mutate problem formulations rather than solutions or instructions, and admit them by inter-solver divergence rather than learning gain, self-consistency, or solver pass rate. These signals reward solver-side progress or confidence rather than the open-endedness of the problem itself. %\alvin{need to explain why these approaches don't work}

\definecolor{berkblue}{HTML}{003262}
\SetCommentSty{berkblueIt}
\newcommand{\berkblueIt}[1]{\textcolor{berkblue}{#1}}
\section{Method}
\label{sec:method}

\begin{algorithm}[t]
\small
\caption{FrontierSmith: Open-Ended Problem Synthesis}
\label{alg:pipeline}
\KwIn{Seed pool $\mathcal{S}$, CP corpus $\mathcal{D}$, LLM $\mathcal{M}$, sample size $B$, solutions per candidate $n$, budgets $N_\text{div}, N_\text{final}$}
\KwOut{Validated open-ended problem set $\mathcal{P}$}
\Indp
$\mathcal{S} \leftarrow \mathcal{S} \cup \mathcal{D}$\tmark*{add CP corpus to seed pool}
$\mathcal{B} \leftarrow$ sample $B$ problems from $\mathcal{S}$\;
$\mathcal{C} \leftarrow \texttt{Mutate}(\mathcal{B}, \mathcal{M})$\tmark*{mutate goals, outputs, inputs}
$\mathcal{C} \leftarrow \texttt{CoarseFilter}(\mathcal{C}, \mathcal{M})$\tmark*{remove non-open-ended}
\ForEach{candidate $c \in \mathcal{C}$}{
    $\{s_1, \ldots, s_n\} \leftarrow$ draw $n$ solutions for $c$ from $\mathcal{M}$\;
    $\hat{d}(c) \leftarrow \frac{1}{\binom{n}{2}} \sum_{i < j} \texttt{LLM-as-a-Judge}(s_i, s_j)$\tmark*{idea divergence}
}
$\mathcal{C} \leftarrow \operatorname{top}_{N_\text{div}}\bigl(\mathcal{C},\; \hat{d}(\cdot)\bigr)$\tmark*{keep highest divergence}
\ForEach{candidate $c \in \mathcal{C}$}{
    $(\mathcal{T}_c, \mathcal{V}_c) \leftarrow \texttt{BuildAndValidate}(c, \mathcal{M})$\tmark*{test cases + verifier with cross-validation}
    Update $\hat{d}(c)$ using score vectors $\mathbf{q}_i = (\mathcal{V}_c(s_i, t_1), \ldots, \mathcal{V}_c(s_i, t_m))$\;
}
$\mathcal{P} \leftarrow \operatorname{top}_{N_\text{final}}\bigl(\mathcal{C},\; \hat{d}(\cdot)\bigr)$\tmark*{re-rank with verified divergence}
$\mathcal{S} \leftarrow \mathcal{S} \cup \mathcal{P}$\tmark*{expand seed pool for next round}
\Indm
\Return{$\mathcal{P}$}
\end{algorithm}

\subsection{Mutation Problem Formulation}
\label{sec:mutation}
%\alvin{what does this mean} %\alvin{I don't get what you meant. Also, we should formally define what `open ended' means} %\alvin{just source? why `missing'?} %\alvin{move?}
Existing work on synthesizing closed-ended coding tasks focuses on constructing test environments rather than problem formulations, because closed-ended formulations can be drawn from existing repositories~\citep{autocode,hardtests,codecontestsplus} and domain-specific datasets~\citep{swesmith,bugpilot}. Open-ended problems have no such source; the formulation itself is the central challenge.

Our key insight is that closed-ended problems can serve as seeds for synthesizing open-ended ones. Targeted mutations to their formulations can remove known optima while preserving a meaningful way to evaluate submission quality.

Concretely, we represent a problem formulation as a tuple $(\mathcal{O}, \mathcal{C}_I, \mathcal{C}_O)$ where $\mathcal{O}$ is the computational goal, $\mathcal{C}_I$ the admissible problem instances, and $\mathcal{C}_O$ the constraints on valid program outputs. In this representation, %\alvin{open ended too right?}
$\mathcal{O}$ can take various forms: a required output, a decision criterion, a property to satisfy, or a quantity to optimize.
We define three mutation types that transform a closed-ended formulation into an open-ended one: %\alvin{need to reference fig 1}

\begin{enumerate}[leftmargin=*]
    \item \textbf{Changing goals} ($\mathcal{O} \to \mathcal{O}'$): replace a decision or exact-answer goal with an optimization-oriented goal that admits graded performance. For example, 2-SAT~\citep{aspvall1979linear} decides whether a Boolean formula with two-literal clauses is satisfiable; a mutation of this goal produces Min-True 2-SAT~\citep{gusfield1992bounded}, which keeps the same input and output but asks for a satisfying assignment that minimizes the number of true variables. %\alvin{you meant to say 2SAT can be replaced with Min2SAT?}

    \item \textbf{Restricting outputs} ($\mathcal{C}_O \to \mathcal{C}_O'$): add or tighten constraints on valid outputs while keeping the underlying goal fixed. For example, the minimum spanning tree problem~\citep{kruskal1956shortest} on a weighted graph asks for a tree connecting all vertices with minimum total edge weight, which admits a greedy solution. Adding per-vertex degree bounds yields the NP-hard degree-constrained spanning tree~\citep{narula1980degree}. At scale, exact solutions become infeasible, and different approximation strategies yield solutions of varying quality. %\alvin{need to explain how this makes the problem open ended}

    \item \textbf{Generalizing inputs} ($\mathcal{C}_I \to \mathcal{C}_I'$): relax structural assumptions on the input domain while keeping the goal and output constraints fixed. For example, the maximum independent set asks for the largest vertex set with no two vertices sharing an edge. On bipartite graphs, it is polynomial via K\H{o}nig's theorem~\citep{konig1990theory}, but on arbitrary graphs it becomes one of Karp's original NP-complete problems~\citep{karp2009reducibility}, similarly making exact solutions infeasible at scale. %\alvin{need to explain how this makes the problem open ended}

\end{enumerate}

In all three mutation types, the resulting problem has no efficient method to certify its optimum at scale and admits a continuous quality measure: both conditions for open-endedness. We implement each mutation by prompting an LLM $\mathcal{M}$ with the seed problem and the desired mutation type; multiple types can apply simultaneously to a single candidate. Given only the problem statement, $\mathcal{M}$ first extracts the original formulation $(\mathcal{O}, \mathcal{C}_I, \mathcal{C}_O)$ and then produces an open-ended candidate $(\mathcal{O}', \mathcal{C}_I', \mathcal{C}_O')$.

\subsection{Problem Filtering}
\label{sec:divergence}
Mutation produces a broad but noisy pool; we apply two filters in sequence to retain only high-quality open-ended candidates. %\alvin{to do what?}

\paragraph{Coarse LLM-as-a-judge filter.}
The first filter is a coarse LLM-as-a-judge check. We prompt $\mathcal{M}$ to check three conditions: the problem defines an optimization objective with no known optimum; multiple distinct strategies are plausible; and a scoring function can meaningfully rank submissions. %\alvin{?} %\alvin{strategies and verifiers are not objectives? I was expecting some description of metrics}
Candidates failing any condition are discarded. %\alvin{move this to the exp section}

%\alvin{I expect the following to be longer description of the 3 criteria mentioned above, but the names don't match}
\paragraph{Idea divergence filter.}
The coarse filter %\alvin{what coarse filter}
removes closed-ended candidates but does not measure solution diversity. If one strategy dominates, the problem effectively degenerates to a closed-ended task. %\alvin{?}
Well-designed open-ended settings exhibit genuine solution diversity. The AtCoder Heuristic Contest~\citep{alebench} and the database join-ordering problem~\citep{steinbrunn1997joinorder} illustrate this: top performers explore fundamentally different ideas rather than refine one dominant approach. This diversity also strengthens RL training: under GRPO~\citep{shao2024deepseekmath}, varied strategies that yield meaningfully different rewards produce a stronger gradient signal than samples following the same heuristic~\citep{ragen2}. We therefore introduce an \emph{idea divergence} filter that directly quantifies this diversity.

\begin{figure}[t]
\centering

% ── Shared style definitions ──
\tikzset{
    fig/.style={scale=1.0, every node/.style={scale=1.0}},
    shapestyle/.style={thick},
    triA/.style={shapestyle, draw=blue!70, fill=blue!15,
                 regular polygon, regular polygon sides=3,
                 minimum size=0.42cm, inner sep=0pt},
    circB/.style={shapestyle, draw=orange!70, fill=orange!15,
                  circle, minimum size=0.42cm, inner sep=0pt},
    rectC/.style={shapestyle, draw=green!60!black, fill=green!15,
                  rectangle, rounded corners=1pt,
                  minimum size=0.38cm, inner sep=0pt},
    starD/.style={shapestyle, draw=red!70, fill=red!15,
                  star, star points=5, minimum size=0.42cm, inner sep=0pt},
}

% ── Top: flow (left-aligned) + shape legend (right-aligned) ──
\begin{tikzpicture}[>={Stealth[length=5pt, width=4pt]}]
    % Flow — left-aligned to match left subfigure
    \def\fx{-4.5}
    \node[draw=black!50, rounded corners=3pt, fill=gray!6,
          font=\footnotesize, inner sep=6pt, anchor=west] (prob) at (\fx, 0) {Problem $c$};
    % Solutions box
    \node[draw=black!40, rounded corners=4pt, fill=white,
          minimum width=2.6cm, minimum height=0.7cm, anchor=west] (solbox) at (\fx+5.2, 0) {};
    % LLM Solver centered between prob and solbox
    \node[draw=black!50, rounded corners=3pt, fill=blue!6,
          font=\footnotesize, inner sep=6pt] (llm) at ($(prob.east)!0.5!(solbox.west)$) {LLM Solvers};
    \draw[->, thick, black] (prob.east) -- (llm.west);
    \draw[->, thick, black] (llm.east) -- (solbox.west);
    \node[triA, scale=0.42]  at (\fx+5.5, 0) {};
    \node[circB, scale=0.42] at (\fx+5.9, 0) {};
    \node[triA, scale=0.42]  at (\fx+6.3, 0) {};
    \node[rectC, scale=0.42] at (\fx+6.7, 0) {};
    \node[circB, scale=0.42] at (\fx+7.1, 0) {};
    \node[starD, scale=0.42] at (\fx+7.5, 0) {};
    \node[font=\footnotesize, text=black] at (solbox.south) [below=2pt] {solutions $\{s_1, \ldots, s_6\}$};

    % Legend — 2×2 grid, aligned with right subfigure
    \def\rx{4}
    \def\ly{0.05}
    \node[triA, scale=0.5]  at (\rx, \ly) {};
    \node[font=\footnotesize, anchor=west] at (\rx+0.25, \ly) {greedy};
    \node[rectC, scale=0.5] at (\rx, \ly-0.45) {};
    \node[font=\footnotesize, anchor=west] at (\rx+0.25, \ly-0.45) {local search};
    \node[circB, scale=0.5] at (\rx+2.0, \ly) {};
    \node[font=\footnotesize, anchor=west] at (\rx+2.25, \ly) {dynamic programming};
    \node[starD, scale=0.5] at (\rx+2.0, \ly-0.45) {};
    \node[font=\footnotesize, anchor=west] at (\rx+2.25, \ly-0.45) {gradient descent};
\end{tikzpicture}

\vspace{4pt}

\begin{subfigure}[t]{0.48\textwidth}
\centering
\begin{tikzpicture}[fig]

% ── Title ──
\node[font=\small\bfseries, text=black] at (1.76, 1.95) {LLM-as-a-judge-based estimation};

% ── 6×6 pairwise matrix (square: 6×0.42 = 2.52cm) ──
\def\sz{0.42}
\def\matx{0.50}
\def\maty{1.10}

% Column labels
\foreach \j/\lab in {0/$s_1$, 1/$s_2$, 2/$s_3$, 3/$s_4$, 4/$s_5$, 5/$s_6$} {
    \node[font=\scriptsize, text=black] at (\matx+\j*\sz+0.5*\sz, \maty+0.25) {\lab};
}
% Row labels with shape glyphs
\foreach \i/\lab/\sh in {0/$s_1$/triA, 1/$s_2$/circB, 2/$s_3$/triA, 3/$s_4$/rectC, 4/$s_5$/circB, 5/$s_6$/starD} {
    \node[font=\scriptsize, text=black, anchor=east] at (\matx-0.40, \maty-\i*\sz-0.5*\sz) {\lab};
    \node[\sh, scale=0.50] at (\matx-0.18, \maty-\i*\sz-0.5*\sz) {};
}

% Diagonal cells
\foreach \i in {0,...,5} {
    \fill[black!12] (\matx+\i*\sz, \maty-\i*\sz) rectangle +(\sz, -\sz);
}
% Same-strategy pairs: (s1,s3) and (s2,s5)
\foreach \r/\c in {0/2, 2/0, 1/4, 4/1} {
    \fill[gray!28] (\matx+\c*\sz, \maty-\r*\sz) rectangle +(\sz, -\sz);
}
% Different-strategy pairs
\foreach \r/\c in {
    0/1, 0/3, 0/4, 0/5,
    1/0, 1/2, 1/3, 1/5,
    2/1, 2/3, 2/4, 2/5,
    3/0, 3/1, 3/2, 3/4, 3/5,
    4/0, 4/2, 4/3, 4/5,
    5/0, 5/1, 5/2, 5/3, 5/4} {
    \fill[blue!22] (\matx+\c*\sz, \maty-\r*\sz) rectangle +(\sz, -\sz);
}
% Grid
\foreach \i in {0,...,6} {
    \draw[thin, black!35] (\matx, \maty-\i*\sz) -- (\matx+6*\sz, \maty-\i*\sz);
}
\foreach \j in {0,...,6} {
    \draw[thin, black!35] (\matx+\j*\sz, \maty) -- (\matx+\j*\sz, \maty-6*\sz);
}

% ── Legend (simple boxes) ──
\fill[blue!22] (\matx+6*\sz+0.25, \maty-1.5*\sz) rectangle +(0.25, -0.25);
\node[font=\footnotesize, text=black, anchor=west] at (\matx+6*\sz+0.55, \maty-1.700*\sz) {different};
\fill[gray!28] (\matx+6*\sz+0.25, \maty-2.3*\sz) rectangle +(0.25, -0.25);
\node[font=\footnotesize, text=black, anchor=west] at (\matx+6*\sz+0.55, \maty-2.575*\sz) {same};

% ── Formula ──
\node[font=\small, text=black] at (1.76, -1.85) {$\hat{d}(c) = \mathrm{avg}\;\texttt{LLM-as-a-Judge}(s_i, s_j)$};

\end{tikzpicture}
\caption*{\textbf{LLM-based estimation.} An LLM judge labels each solution pair as same- or different-strategy. $\hat{d}(c)$ is the fraction of pairs judged different. Blue cells: different-strategy; gray cells: same-strategy.}
\end{subfigure}
\hfill
\begin{subfigure}[t]{0.48\textwidth}
\centering
\begin{tikzpicture}[fig]

% ── Title ──
\node[font=\small\bfseries, text=black] at (1.76, 1.95) {Execution-based estimation};

% ── 6×6 heatmap (square: 6×0.42 = 2.52cm), col 4 = ⋯ ──
\def\hx{0.50}
\def\hy{1.10}
\def\cw{0.42}
\def\ch{0.42}

% Column labels: t1, t2, t3, ⋯, t_m
\foreach \j/\lab in {0/$t_1$, 1/$t_2$, 2/$t_3$, 3/$\cdots$, 4/{}, 5/$t_m$} {
    \node[font=\scriptsize, text=black] at (\hx+\j*\cw+0.5*\cw, \hy+0.25) {\lab};
}
% Row labels with shape glyphs
\foreach \i/\lab/\sh in {0/$s_1$/triA, 1/$s_2$/circB, 2/$s_3$/triA, 3/$s_4$/rectC, 4/$s_5$/circB, 5/$s_6$/starD} {
    \node[font=\scriptsize, text=black, anchor=east] at (\hx-0.40, \hy-\i*\ch-0.5*\ch) {\lab};
    \node[\sh, scale=0.50] at (\hx-0.18, \hy-\i*\ch-0.5*\ch) {};
}

% Heatmap fills (green gradient). Same-strategy rows have similar patterns.
\foreach \j/\v in {0/70, 1/55, 2/18, 3/30, 4/42, 5/30} {
    \fill[green!\v] (\hx+\j*\cw, \hy-0*\ch) rectangle +(\cw,-\ch);}
\foreach \j/\v in {0/8, 1/18, 2/70, 3/30, 4/8, 5/18} {
    \fill[green!\v] (\hx+\j*\cw, \hy-1*\ch) rectangle +(\cw,-\ch);}
\foreach \j/\v in {0/55, 1/70, 2/18, 3/30, 4/42, 5/30} {
    \fill[green!\v] (\hx+\j*\cw, \hy-2*\ch) rectangle +(\cw,-\ch);}
\foreach \j/\v in {0/30, 1/30, 2/42, 3/30, 4/70, 5/55} {
    \fill[green!\v] (\hx+\j*\cw, \hy-3*\ch) rectangle +(\cw,-\ch);}
\foreach \j/\v in {0/18, 1/8, 2/55, 3/30, 4/18, 5/18} {
    \fill[green!\v] (\hx+\j*\cw, \hy-4*\ch) rectangle +(\cw,-\ch);}
\foreach \j/\v in {0/8, 1/8, 2/18, 3/30, 4/42, 5/70} {
    \fill[green!\v] (\hx+\j*\cw, \hy-5*\ch) rectangle +(\cw,-\ch);}

% Grid
\foreach \i in {0,...,6} {
    \draw[thin, black!35] (\hx, \hy-\i*\ch) -- (\hx+6*\cw, \hy-\i*\ch);
}
\foreach \j in {0,...,6} {
    \draw[thin, black!35] (\hx+\j*\cw, \hy) -- (\hx+\j*\cw, \hy-6*\ch);
}

% ── Colorbar (left of heatmap, 0 to 1, light to dark green) ──
\pgfmathsetmacro{\cbx}{\hx - 1.05}
\pgfmathsetmacro{\cbtop}{\hy}
\def\cbw{0.18}
\pgfmathsetmacro{\cbh}{6*\ch}
\foreach \k in {0,1,...,30} {
    \pgfmathsetmacro{\frac}{\k/30}
    \pgfmathsetmacro{\yy}{\cbtop - \frac*\cbh}
    \pgfmathsetmacro{\gv}{70*(1-\frac)}
    \fill[green!\gv] (\cbx, \yy) rectangle +(\cbw, -\cbh/30);
}
\draw[thin, black!40] (\cbx, \cbtop) rectangle +(\cbw, -\cbh);
\node[font=\tiny, text=black, anchor=east] at (\cbx-0.04, \cbtop) {$1$};
\node[font=\tiny, text=black, anchor=east] at (\cbx-0.04, \cbtop-\cbh) {$0$};
\node[font=\footnotesize, text=black, rotate=90, anchor=south] at (\cbx-0.08, \cbtop-\cbh/2) {test case score};

% ── Score vector brace + labels ──
\pgfmathsetmacro{\brx}{\hx+6*\cw+0.06}
\draw[decorate, decoration={brace, amplitude=3pt}, semithick, black!60]
    (\brx, \hy) -- (\brx, \hy-6*\ch);
\foreach \i in {0,...,5} {
    \pgfmathsetmacro{\yi}{\hy - \i*\ch - 0.5*\ch}
    \node[font=\footnotesize, text=black, anchor=west] at (\brx+0.14, \yi)
        {$\mathbf{q}_{\the\numexpr\i+1}$};
}

% ── Formula ──
\node[font=\small, text=black] at (1.76, -1.85)
    {$\hat{d}(c) = \mathrm{avg}\;\frac{1}{\sqrt{m}}\lVert\mathbf{q}_i - \mathbf{q}_j\rVert_2$};

\end{tikzpicture}
\caption*{\textbf{Execution-grounded estimation.} Each solution is run on the test cases $t_1, \ldots, t_m$ and scored by the verifier, yielding a score vector $\mathbf{q}_i$. $\hat{d}(c)$ is the average pairwise distance between score vectors.}
\end{subfigure}

\caption{Estimating idea divergence $\hat{d}(c)$, the probability that two independently sampled LLM-generated solutions to candidate problem $c$ use different algorithmic strategies. We draw 6 solutions $s_1, \ldots, s_6$ from an LLM solver and estimate $\hat{d}(c)$ in two complementary ways. The shape attached to each $s_i$ denotes its underlying algorithmic strategy.}
\label{fig:divergence}
\end{figure}

\paragraph{Definition.}
For a candidate problem $c = (\mathcal{O}, \mathcal{C}_I, \mathcal{C}_O)$, we define idea divergence as the probability that two independently generated solutions use different algorithmic strategies (Figure~\ref{fig:divergence}):
\begin{equation}
    d(c) \coloneqq \mathbb{P}_{s_i, s_j \sim \text{Solver}(c)}\bigl[\text{strategy}(s_i) \neq \text{strategy}(s_j)\bigr].
    \label{eq:divergence}
\end{equation}
$\text{Solver}(c)$ is the distribution over LLM-generated solutions for $c$, and $\text{strategy}(\cdot)$ maps each solution to its core algorithmic idea.
Computing $d(c)$ exactly is intractable; we provide two complementary estimates below.
%\alvin{move this to before this para an say that we have two estimates for d(c)}

\paragraph{LLM-based estimate.}
We draw $n$ independent solutions $s_1, \ldots, s_n \sim \text{Solver}(c)$. An LLM-as-a-judge labels each pair $(s_i, s_j)$ as same- or different-strategy:
\begin{equation}
    \hat{d}(c) \coloneqq \frac{1}{\binom{n}{2}} \sum_{i < j} \texttt{LLM-as-a-Judge}\bigl(s_i, s_j\bigr).
    \label{eq:divergence_estimate}
\end{equation}
where each judge call returns 1 if the strategies differ and 0 otherwise.
A naive implementation requires $O(n^2)$ judge calls. We batch the calls instead, scoring all pairs in a small group per query and averaging across groups. %\alvin{LLM-as-a-judge returns 0/1?}

\paragraph{Execution-grounded estimate.}
After testing infrastructure (\S\ref{sec:testing}), we complement with an execution-based estimate. Given test cases $\mathcal{T}_c = \{t_1, \ldots, t_m\}$ and verifier $\mathcal{V}_c$ for $c$, the score vector $\mathbf{q}_i = (\mathcal{V}_c(s_i, t_1), \ldots, \mathcal{V}_c(s_i, t_m))$ records $s_i$'s per-test-case performance, and we estimate divergence as
\begin{equation}
    \hat{d}(c) \coloneqq \frac{1}{\binom{n}{2}} \sum_{i < j} \frac{1}{\sqrt{m}} \left\lVert \mathbf{q}_i - \mathbf{q}_j \right\rVert_2.
\end{equation}

\paragraph{Selection.}
The two estimates above are complementary. The LLM-based estimate captures semantic differences in algorithmic strategy, e.g., greedy vs.\ dynamic programming. The execution-based estimate captures behavioral differences across test cases, e.g., two similar solutions that trade speed against accuracy. Together they form a two-stage funnel (Algorithm~\ref{alg:pipeline}): the LLM-based estimate is applied first since it requires no test infrastructure, retaining the top $N_\text{div}$ candidates. After testing infrastructure (\S\ref{sec:testing}), the execution-based estimate refines the ranking, and the top $N_\text{final}$ are selected as the final problem set $\mathcal{P}$. 

\subsection{Testing Infrastructure}
\label{sec:testing}

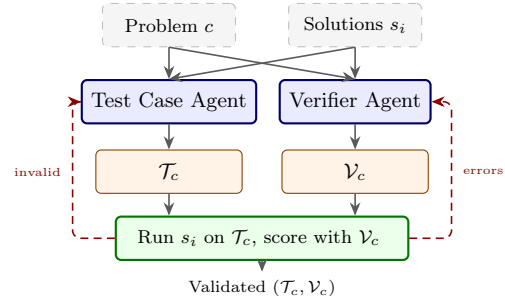
\begin{wrapfigure}{r}{0.55\columnwidth}
\vspace{-8pt}
\centering
\begin{tikzpicture}[
    >={Stealth[length=5pt, width=4pt]},
    scale=0.88, every node/.style={scale=0.88},
    box/.style={
        draw, rounded corners=2pt, minimum height=0.65cm,
        align=center, font=\small, inner sep=4pt
    },
    agent/.style={
        box, fill=blue!8, draw=blue!40!black, thick,
        minimum width=2.2cm
    },
    artifact/.style={
        box, fill=orange!10, draw=orange!50!black,
        minimum width=2.2cm
    },
    input/.style={
        box, fill=gray!8, draw=gray!50, dashed,
        minimum width=2.0cm, font=\footnotesize
    },
    validate/.style={
        box, fill=green!8, draw=green!45!black, thick,
        minimum width=4.4cm, font=\footnotesize
    },
    arr/.style={->, semithick, color=black!65},
    feedarr/.style={->, semithick, color=red!55!black, densely dashed},
    lbl/.style={font=\tiny, text=red!55!black},
]

% Two columns, centered, with margins for feedback arrows
\def\colL{1.6}
\def\colR{4.4}
\def\colMid{3.0}

% Bounding box with margins for feedback
\useasboundingbox (0, 0.35) rectangle (6.0, -4.55);

% === Inputs ===
\node[input] (prob) at (\colL, 0) {Problem $c$};
\node[input] (sols) at (\colR, 0) {Solutions $s_i$};

% === Agents ===
\node[agent] (tcagent) at (\colL, -1.15) {Test Case Agent};
\node[agent] (vagent)  at (\colR, -1.15) {Verifier Agent};

% === Artifacts ===
\node[artifact] (tc) at (\colL, -2.2) {$\mathcal{T}_c$};
\node[artifact] (vf) at (\colR, -2.2) {$\mathcal{V}_c$};

% === Cross-validation ===
\node[validate] (xval) at (\colMid, -3.2) {
    Run $s_i$ on $\mathcal{T}_c$, score with $\mathcal{V}_c$
};

% === Output ===
\node[font=\scriptsize] (out) at (\colMid, -3.95) {
    Validated $(\mathcal{T}_c, \mathcal{V}_c)$
};

% === Downward flow ===
\draw[arr] (prob) -- (tcagent);
\draw[arr] (sols) -- (vagent);
% Cross arrows: straight diagonal lines
\draw[arr] (prob.south) -- (vagent.north);
\draw[arr] (sols.south) -- (tcagent.north);
\draw[arr] (tcagent) -- (tc);
\draw[arr] (vagent)  -- (vf);
\draw[arr] (tc.south) -- (tc.south |- xval.north);
\draw[arr] (vf.south) -- (vf.south |- xval.north);
\draw[arr] (xval) -- (out);

% === Feedback arrows — well outside the boxes ===
% Left: xval → TC agent
\draw[feedarr, rounded corners=3pt]
    (xval.west) -- (0.15, -3.2) -- (0.15, -1.15) -- (tcagent.west);
\node[lbl, anchor=east] at (0.1, -2.2) {invalid};

% Right: xval → Verifier agent
\draw[feedarr, rounded corners=3pt]
    (xval.east) -- (5.85, -3.2) -- (5.85, -1.15) -- (vagent.east);
\node[lbl, anchor=west] at (5.9, -2.2) {errors};

\end{tikzpicture}
\vspace{-8pt}
\caption{Test case and verifier synthesis. A test case agent produces inputs $\mathcal{T}_c$; a verifier agent produces a scoring program $\mathcal{V}_c$. Sampled solutions are run on both; each agent uses the other's output to validate its own (dashed), iterating until consistent.}
\label{fig:test-infra}
% \vspace{-18pt}
\end{wrapfigure}

\paragraph{Test case generation.}
For each surviving candidate $c$, a test case agent generates a set of inputs $\mathcal{T}_c$ (Figure~\ref{fig:test-infra}).
Closed-ended test cases target corner cases and boundary conditions; open-ended test cases must stress-test different algorithmic strategies.
Following prior work on test synthesis~\citep{autocode, codecontestsplus}, we prompt the agent to write test-case generator programs %\alvin{test}
that produce inputs of varying size %\alvin{what does scale mean}
and structure, such as sparse versus dense graphs or uniform versus skewed distributions.
The test-case agent %\alvin{which}
also receives the solutions sampled by the divergence estimate of \S\ref{sec:divergence}, allowing it to craft adversarial inputs that expose where specific strategies break down.
\paragraph{Verifier generation.}
Each candidate's objective $\mathcal{O}$ orders solutions but yields no bounded score. A separate verifier agent translates $\mathcal{O}$ into a scoring program $\mathcal{V}_c$ that returns a normalized score in $[0, 1]$.
One common approach is to normalize against a baseline solution $s^*$: $\mathcal{V}_c$ returns $0$ when $s$ fails to improve on $s^*$ and a value in $[0, 1)$ otherwise, scaling with the size of the improvement. The baseline need not be strong; even a greedy heuristic or random valid solution suffices. %\alvin{how do you ensure that the generated solution is weak enough}
Note that we can ensure $\mathcal{O}(\cdot, t) > 0$ by adding a constant offset.
We assume $\mathcal{O}(\cdot, t) > 0$. Letting $\sigma_{\mathcal{O}} = +1$ for maximization objectives and $-1$ for minimization,
\begin{equation}
    \mathcal{V}_c(s, t) = \max\!\left(0,\;\sigma_{\mathcal{O}}\cdot\frac{\mathcal{O}(s, t) - \mathcal{O}(s^*, t)}{\max\{\mathcal{O}(s, t),\,\mathcal{O}(s^*, t)\}}\right).
\end{equation}
Solutions that crash, time out, or produce unparseable output receive a score of 0.

\paragraph{Cross-validation protocol.}
The test case agent and the verifier agent use each other's output for self-validation.
The test case agent runs the sampled solutions through the verifier: if a solution crashes or produces unparseable output on a test input, that input is invalid, and the test-case agent %\alvin{which}
revises it.
The verifier agent scores the sampled solutions from \S\ref{sec:divergence} %\alvin{what `solutions' are you referring to}
on the generated test cases: if scores collapse to a narrow band or contradict the relative quality of solutions, the scoring program is flawed, and the agent revises it.
Each agent iterates until the other's output no longer exposes errors in its own; candidates that fail to converge within a fixed number of rounds are discarded. %\alvin{how do you know when to stop?}
In our experiments, 10\% of candidates that enter this stage produce a validated $(\mathcal{T}_c, \mathcal{V}_c)$ pair, which then feeds the execution-based idea-divergence re-ranking of \S\ref{sec:divergence}.

\section{Experiments}
\label{sec:experiments}
\subsection{Experimental Setup}

\paragraph{Benchmarks.}
We evaluate on two open-ended coding benchmarks.
FrontierCS~\citep{frontiercs} contains 240 open-ended problems across two tracks. We use the 172 algorithmic problems, which require only local computation; the remaining 68 problems require cloud infrastructure (e.g., GPUs, external services) and are excluded. Submissions are scored on a continuous 0--100 scale.
ALE-bench~\citep{alebench} draws tasks from AtCoder Heuristic Contests and evaluates submissions using performance-based ratings. %\alvin{this para should start with `We use 2 benchmarks: ...'}
We use the ALE-bench-lite subset (10 tasks) for evaluation.

\paragraph{Baselines.}
We compare five configurations:
(1)~Base: the pretrained model without RL;
(2)~FrontierCS: RL on 172 human-curated FrontierCS algorithmic problems;
(3)~ALE-bench:  RL on 40 ALE-bench tasks, with rewards computed from each problem's public test set.
(4)~HardTests: RL on 200 problems randomly sampled from the 47{,}136 closed-ended competitive programming problems from HardTests~\citep{hardtests}, trained with binary rewards (1 if all test cases pass, 0 otherwise), following the standard competitive programming setup~\citep{codeforces, icpc};
(5)~Random Reward: RL on 172 FrontierCS algorithmic problems with the raw rewards drawn from $\mathcal{U}[0, 100]$. Following~\citet{shao2025spurious}, this is a spurious reward %\alvin{why is there a -}
control: it tests whether gains require task-specific reward signals or arise from RL dynamics under uninformative rewards.

\paragraph{Problem synthesis.}
We run Algorithm~\ref{alg:pipeline} for 4 iterations. Each iteration samples $B=1{,}000$ seed problems from the pool (initialized with HardTests~\citep{hardtests}) with $N_{\text{div}}=100$ and $N_{\text{final}}=50$, yielding 50 synthetic problems.
We use GPT-5.4 Thinking~\citep{openai2026gpt54} with medium thinking effort for problem formulation mutation (\S\ref{sec:mutation}), the coarse LLM-as-a-judge filter, the LLM-based divergence estimate (\S\ref{sec:divergence}), and Claude Sonnet 4.6~\citep{anthropic2026sonnet46} with default thinking effort for sampling solutions ($n=10$ per candidate), test case generation, and verifier generation (\S\ref{sec:testing}).
The generated problems follow the FrontierCS~\citep{frontiercs} format, and we reuse its evaluation sandbox %\alvin{judger?}
to execute code and run verifiers.

\paragraph{Training configurations.}
We use veRL~\citep{sheng2024hybridflow} with GRPO, learning rate is set to $10^{-6}$ for Qwen3.5-9B and $5 \times 10^{-7}$ for Qwen3.5-27B, rollout batch size $8$, and group size $8$. We train for 100 steps and evaluate every 10 steps. Qwen3.5-9B trains on 8 A100 GPUs for 1.5 days with maximum response length $16{,}000$ tokens; Qwen3.5-27B trains on 32 H200 GPUs for 1.5 days with maximum response length $32{,}000$ tokens. For the 27B model, we only report Base, FrontierCS, HardTests, and FrontierSmith due to our computing budget.

\subsection{Main Results}

\begin{table}[t]
\centering
\renewcommand{\arraystretch}{1.3}
\resizebox{0.9\textwidth}{!}{%
\begin{tabular}{l cccc cccc}
\toprule
& \multicolumn{4}{c}{\textbf{Qwen3.5-9B}} & \multicolumn{4}{c}{\textbf{Qwen3.5-27B}} \\
\cmidrule(lr){2-5} \cmidrule(lr){6-9}
& \multicolumn{2}{c}{FrontierCS} & \multicolumn{2}{c}{ALE-bench} & \multicolumn{2}{c}{FrontierCS} & \multicolumn{2}{c}{ALE-bench} \\
\cmidrule(lr){2-3} \cmidrule(lr){4-5} \cmidrule(lr){6-7} \cmidrule(lr){8-9}
\textbf{Training Data} & Avg@5 & Best@5 & Avg@5 & Best@5 & Avg@5 & Best@5 & Avg@5 & Best@5 \\
\midrule
Base        & 1.80 &  5.00 & 327.22 & 327.22 & 7.70 & 13.91 & 352.52 & 498.00 \\
\midrule
FrontierCS (172) & \cellcolor{lightred}\textbf{11.17} & \cellcolor{lightred}\textbf{16.29} & 558.49 & \underline{762.28} & \cellcolor{lightred}\underline{13.98} & \cellcolor{lightred}\underline{21.92} & \underline{543.80} & 843.80 \\
ALE-bench (40) & \underline{10.64} & 15.70 & \cellcolor{lightred}\textbf{657.40} & \cellcolor{lightred}750.15 & - & - & - & -\\
\midrule
\rowcolor{lightblue}
FrontierSmith (200)    & 10.62 & \underline{15.73} & \underline{633.58} & \textbf{782.30} & \textbf{19.82} & \textbf{29.38}  & \textbf{661.64} & \textbf{938.10} \\
HardTests (200)  & 5.38 & 11.19 & 397.18 & 512.83 & 11.20 & 18.37 & 529.12 & \underline{922.50} \\
Random Reward (172)      & 3.04 & 6.93 & 376.82 & 466.73 & - & - & - & - \\
\bottomrule
\end{tabular}%
}
\vspace{2ex}
\caption{Performance on FrontierCS and ALE-bench. We train for 100 steps and evaluate every 10 steps; reported numbers are the best checkpoint for each metric. Avg@5 and Best@5: average and maximum score over 5 samples per problem. Parentheses: number of training problems. \textbf{Bold}: best; \underline{underline}: second best. Red cells: training data overlaps with the evaluation benchmark.}
\vspace{-2ex}
\label{tab:main-results}
\end{table}

Table~\ref{tab:main-results} reports the main results. FrontierSmith achieves competitive or superior performance to human-curated training data on both benchmarks, and the gains hold across model sizes (9B and 27B), suggesting that our synthesis pipeline scales with model capacity.

\paragraph{FrontierSmith vs.\ human-curated open-ended data.}
On Qwen3.5-9B, FrontierSmith achieves \nBOursFC{} Avg@5 on FrontierCS, close to \nBHumanFC{} from human-curated FrontierCS problems, a gap of only \nBGapHumanFC{} points. On ALE-bench, FrontierSmith achieves the highest Best@5 (782.30) across all configurations, and its Avg@5 (\nBOursALE{}) surpasses the FrontierCS-trained baseline (\nBHumanALE{}), trailing only the ALE-bench-trained model (657.40) which benefits from in-domain data. On Qwen3.5-27B, FrontierSmith outperforms the human-curated baseline on both benchmarks: FrontierCS (\tBOursFC{} vs.\ \tBHumanFC{}) and ALE-bench (\tBOursALE{} vs.\ \tBHumanALE{}), improving from a base of \tBBaseFC{} and \tBBaseALE{}. These results suggest that automated synthesis can largely substitute for expensive human curation, with strong cross-benchmark generalization.

\paragraph{FrontierSmith vs.\ closed-ended data.} 
% \alvin{I think you meant training on open ended vs closed ended problems}
HardTests problems serve as the seed corpus for FrontierSmith's mutations, yet training directly on them yields only \nBHardFC{} on FrontierCS and \nBHardALE{} on ALE-bench for the 9B model, well below FrontierSmith's \nBOursFC{} and \nBOursALE{}. The same pattern holds for the 27B model (\tBHardFC{} vs.\ \tBOursFC{} on FrontierCS). This demonstrates that mutation transforms closed-ended problems into more effective open-ended training data.

\paragraph{Random reward control.}
The random reward baseline scores \nBRandFC{} on FrontierCS and \nBRandALE{} on ALE-bench, which is similar to the untrained base (\nBBaseFC{} and \nBBaseALE{}). This rules out problem-format exposure as the source of gains, consistent with~\citet{shao2025spurious}.

\subsection{Filter Analysis}

\begin{figure}[t]
\centering
\includegraphics[width=0.91\linewidth]{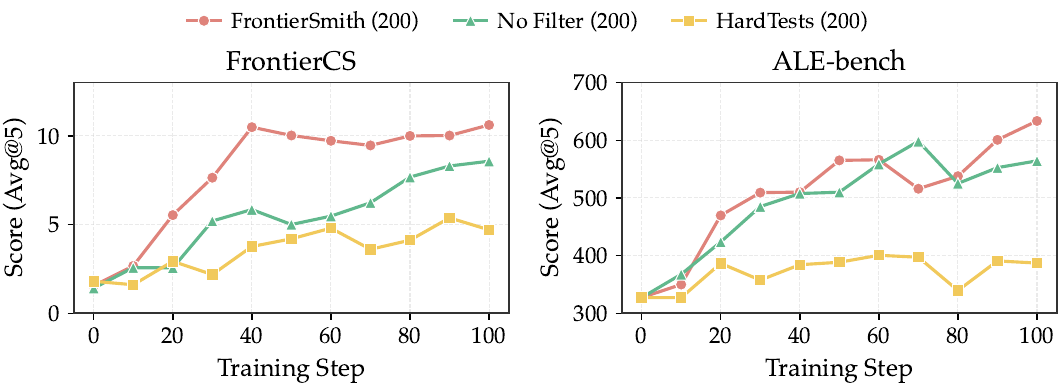}
\caption{Test performance over training steps on FrontierCS and ALE-bench (Qwen3.5-9B, 200 problems each). FrontierSmith and HardTests are the same runs as Table~\ref{tab:main-results}; ``No Filter'' skips both filters. FrontierSmith consistently outperforms the no-filter ablation and the HardTests baseline.}
\label{fig:nofilter-ablation}
% \vspace{-2ex}
\end{figure}

Figure~\ref{fig:nofilter-ablation} shows test performance over training steps for three 200-problem configurations: FrontierSmith and HardTests are the same runs as in Table~\ref{tab:main-results}; No Filter skips both filters and directly builds test infrastructure for 200 mutated problems. 
On FrontierCS, FrontierSmith reaches 10.62 Avg@5, compared with 8.57 for the no-filter variant. The same trend holds on ALE-bench, where FrontierSmith achieves 633.6 versus 564.4. These results suggest that idea-divergence filtering improves both in-domain performance and cross-benchmark generalization.
Both filtered and unfiltered synthetic data outperform the closed-ended HardTests baseline (5.38 and 397.2).
We next validate the two filters individually.

\paragraph{Coarse filter validation.}
To validate the coarse LLM-as-a-judge filter (\S\ref{sec:divergence}), we apply it to 100 closed-ended competitive programming problems from HardTests~\citep{hardtests} using GPT-5.4 Thinking~\citep{openai2026gpt54} as the judge. Since all problems in this set are closed-ended, any retained candidate is a false positive. The filter rejects 91 and retains 9, yielding a false-positive rate of 9\%.
Conversely, applying the filter to 100 FrontierCS problems rejects 19, a false-negative rate of 19\%. This is acceptable because the filter is designed to preserve high-quality open-ended problems rather than maximizing recall.

\begin{figure*}[t]
\centering
\begin{subfigure}[t]{0.40\textwidth}
\centering
\includegraphics[width=\linewidth]{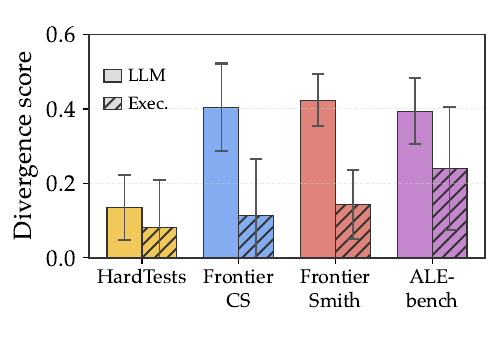}
\end{subfigure}
\hfill
\begin{subfigure}[t]{0.57\textwidth}
\centering
\includegraphics[width=\linewidth]{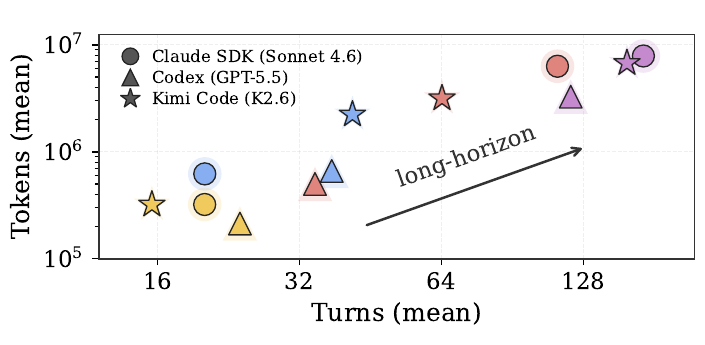}
\end{subfigure}
\caption{Comparison of four problem sources. Color encodes the problem source in both panels. \textbf{Left:} Idea divergence cleanly separates open-ended from closed-ended problems. \textbf{Right:} Points show source-agent geometric means of turns and tokens; FrontierSmith and human-curated open-ended problems elicit increasing horizon agent behavior.}
% \alvin{what's the color code of the right plot? instead of `long horizon' I suggest `increasing horizon'}}
\label{fig:analysis}
\vspace{-2ex}
\end{figure*}

\paragraph{Idea divergence as a classifier.}
\label{sec:divergence-classifier}
To test whether idea divergence separates open-ended from closed-ended problems, we sample 10 tasks from each of four sources: FrontierCS (human-curated), FrontierSmith (synthetic), HardTests (closed-ended, sampled from the hardest difficulty tier in human competitive programming contests), and ALE-bench. For each task, we sample $n{=}10$ solutions using Claude Sonnet 4.6 and compute the two divergence estimates (\S\ref{sec:divergence}). 

The left panel of Figure~\ref{fig:analysis} shows the results. Using the LLM-based estimate, the three open-ended sources (FrontierCS, FrontierSmith, and ALE-bench) all score around 0.4, roughly $3\times$ higher than HardTests (0.14). FrontierSmith scores slightly higher than the human-curated FrontierCS (0.42 vs.\ 0.40), suggesting that our filtering pipeline selects problems with genuine solution diversity comparable to the human-curated ones.
% \alvin{is this actually statistically significant?} 
The execution-grounded estimate shows a similar trend: HardTests remains lowest (0.08), while FrontierCS, FrontierSmith, and ALE-bench score 0.11, 0.14, and 0.24, respectively. Although the margin between HardTests and the open-ended sources is smaller, the separation remains clear.

\subsection{Long-Horizon Code Agent Behavior}

Open-ended problems often lead to extended interactions: agents benefit from iterating on solutions, running tests, and refining strategies over many turns. Using the same 40 tasks from \S\ref{sec:divergence-classifier}, we run three code agents via the Harbor evaluation framework~\citep{harbor2026}: Claude SDK with Sonnet 4.6~\citep{anthropic2026sonnet46}, Codex with GPT-5.5~\citep{openai2026gpt55}, and Kimi Code with K2.6~\citep{kimik26}. The right panel of Figure~\ref{fig:analysis} plots geometric-mean turns and tokens for each source-agent pair. ALE-bench drives all three agents into the long-horizon regime, where every pair exceeds 100 turns and $3{\times}10^6$ tokens. FrontierSmith exhibits the same long-horizon behavior, with Claude SDK reaching 113 turns and $6.3{\times}10^6$ tokens, on par with ALE-bench. HardTests and FrontierCS stay in the short-horizon regime.

\section{Discussion}
\label{sec:discussion}

\paragraph{Limitations.}
Our current pipeline operates entirely within self-contained algorithmic environments: each problem takes a text input and produces a program output, with no external dependencies. This excludes repo-level open-ended tasks that require complex environment setup, such as system optimization with cloud infrastructure, GPU kernel tuning~\citep{cao2026k}, or multi-file software engineering. Extending FrontierSmith to these settings would require generating not only problem formulations and verifiers but also reproducible execution environments.
Additionally, due to compute budget constraints, our RL training is limited to 100 steps of single-turn GRPO. We do not explore agentic RL and leave this as future work, where the policy interacts with the environment over multiple turns, which is a natural next step given the long-horizon nature of the problems we synthesize.

\section{Conclusion}
\label{sec:conclusion}

We introduced FrontierSmith, an automated pipeline that transforms closed-ended coding problems into open-ended ones using targeted mutations and idea-divergence filtering. Training Qwen3.5-9B and 27B on FrontierSmith-generated problems achieves performance competitive with training on human-curated open-ended data, while substantially outperforming training on closed-ended problems and generalizing strongly across benchmarks. Our analysis further shows that idea divergence serves as a reliable signal of open-endedness, and that synthesized problems elicit long-horizon agent behavior comparable to human-curated tasks, with agents spending substantially more turns and tokens than on closed-ended problems. Together, these findings show that open-ended coding data can be synthesized at scale without expensive expert curation.
FrontierSmith therefore provides a scalable source of training data for RL on long-horizon, open-ended coding tasks.

\section*{Acknowledgments}
We thank Youliang Yuan, Ruyi Ji, Zhifei Li, the Harbor team, the Modal team, and the Laude Institute. We also thank the Kimi team for providing access to the Kimi Code API.

{\clearpage} % prevent wrapfigure bleed into bibliography
\bibliographystyle{plainnat}
\bibliography{references}

\clearpage
\beginappendix
\section{Example Synthesized Problems}
\label{app:example-problem}

Below are two example open-ended problems produced by FrontierSmith, each mutated from a different closed-ended seed.

\subsection{Concat Factory Compression Challenge}

Polycarp's toy language has only two commands:
\begin{enumerate}[leftmargin=*]
    \item \textbf{Literal creation} (\texttt{L s}): Creates a new variable whose value is the lowercase string \texttt{s} (length 1--8).
    \item \textbf{Concatenation} (\texttt{C a b}): Creates a new variable whose value is the concatenation of the values of previously created variables \texttt{a} and \texttt{b}.
\end{enumerate}
Variables are numbered automatically in creation order, starting from 1. For each test case, you are given a list of target strings. Your task is to output a valid program that constructs every target string in some variable, while keeping the program as small as possible.

\subsubsection*{Input}
The first line contains an integer $T$ (number of test cases). For each test case: the first line contains $q$ (number of targets), followed by $q$ lines each containing one target string.

\subsubsection*{Output}
For each test case, output: (1) an integer $m$ (number of commands); (2) $m$ lines, each \texttt{L s} or \texttt{C a b}; (3) one line with $q$ integers $r_1, \ldots, r_q$, where $r_i$ is the index of a variable whose value equals the $i$-th target.

\subsubsection*{Feasibility}
A solution is feasible if $1 \le m \le 5000$, every command has valid syntax, every literal has length 1--8 with only lowercase letters, every \texttt{C a b} references previously created variables, every variable's value has length at most the maximum target length, and for every $i$, $\texttt{value}[r_i] = t_i$.

\subsubsection*{Objective}
For a feasible solution, the cost is
\[
C = m + \sum_{\text{literal commands } L\,s} |s|.
\]
The baseline cost $B$ for each test case is computed by splitting each target into blocks of length at most 8 without reuse:
\[
B = \sum_{\text{targets } t} \left(|t| + 2\left\lceil \frac{|t|}{8} \right\rceil - 1\right).
\]
The test-case score is $S = 10^6 \cdot \min(2,\; B/C)$. The final score is the mean of $S$ over all test cases.

\subsubsection*{Constraints}
$1 \le T \le 20$, \; $1 \le q \le 200$, \; $1 \le |t_i| \le 400$, \; sum of target lengths per test case $\le 40{,}000$. Time limit: 3\,s. Memory limit: 512\,MB.

\subsubsection*{Example}
Input: 3 targets \texttt{haha}, \texttt{ahaha}, \texttt{hahahaha}.

Program: \texttt{L ha}, \texttt{L a}, \texttt{C 1 1}, \texttt{C 2 3}, \texttt{C 3 3}. Variables: $1{=}$\texttt{ha}, $2{=}$\texttt{a}, $3{=}$\texttt{haha}, $4{=}$\texttt{ahaha}, $5{=}$\texttt{hahahaha}. Targets matched by variables 3, 4, 5.

Cost $C = 5 + (2+1) = 8$. Baseline $B = 5 + 6 + 9 = 20$. Score $= 10^6 \cdot \min(2, 20/8) = 2 \times 10^6$ (cap).

\subsection{Yae Village Defense Network}

The ancient village of Yae has $n$ houses, numbered $1 \ldots n$. House 1 is the Grand Shrine, where Sakura starts and ends every day. There are $m$ candidate bidirectional roads; road $i$ connects houses $u_i, v_i$ with build cost $c_i$ and travel time $d_i$.

Before day 1, Sakura may build a subset of roads with total cost $\le B$. Over $D$ days, Collapse beasts attack: attack $j$ targets house $h_j$, is active on days $[a_j, b_j]$, and causes damage $w_j$ if not cleared. On each day $t$, Sakura starts at house 1, walks along built roads (total travel time $\le L_t$), and must return to house 1. Visiting a house clears all active attacks there.

The task is to choose which roads to build and one patrol per day to \textbf{minimize total damage} from uncleared attacks.

\subsubsection*{Input}
One instance: $n$, $m$, $D$, $A$ (number of attacks), $B$ (budget), followed by $m$ road descriptions $(u_i, v_i, c_i, d_i)$, daily time limits $L_1, \ldots, L_D$, and $A$ attack descriptions $(h_j, a_j, b_j, w_j)$.

\subsubsection*{Output}
Line 1: $k$ built road indices. Lines $2$ to $D+1$: for each day, a sequence of built road indices forming a walk from house 1 back to house 1.

\subsubsection*{Feasibility}
Built roads must have total cost $\le B$; each daily patrol must use only built roads, follow valid adjacencies, start and end at house 1, and respect the daily time limit $L_t$.

\subsubsection*{Objective}
Minimize $\text{Damage} = \sum_{j \text{ not cleared}} w_j$. The baseline builds no roads and stays at house 1 every day, clearing only attacks at house 1. For baseline damage $B_{\text{dmg}}$ and your damage $D_{\text{your}}$:
\[
\text{Score} = \lfloor 10^6 \cdot \min(10,\; (B_{\text{dmg}} + 1) / (D_{\text{your}} + 1)) \rfloor.
\]
Final score is the mean over all test files.

\subsubsection*{Constraints}
$2 \le n \le 400$, \; $1 \le m \le 3000$, \; $1 \le D \le 60$, \; $1 \le A \le 4000$, \; $1 \le c_i, d_i \le 10^6$, \; $1 \le B \le 10^9$, \; $0 \le L_t \le 10^9$, \; $1 \le w_j \le 10^{12}$. Time limit: 3\,s. Memory limit: 256\,MB.

\subsubsection*{Example}
4 houses, 4 candidate roads, 3 days, 4 attacks, budget 4. Building roads 1--2 and 2--3 (cost 4) and patrolling: day 1 visits house 2 (clears attack, energy 5), day 2 visits house 3 (clears attack, energy 8), day 3 stays home (clears house-1 attack, energy 4). The attack at house 4 is missed: total damage $= 6$, baseline damage $= 19$, score $= \lfloor 10^6 \cdot (19{+}1)/(6{+}1) \rfloor = 2{,}857{,}142$.

\end{document}